\documentclass[11pt]{article}

\usepackage[final]{acl}

\usepackage{times}
\usepackage{latexsym}
\usepackage[T1]{fontenc}
\usepackage[utf8]{inputenc}
\usepackage{microtype}
\usepackage{inconsolata}
\usepackage{graphicx}
\usepackage{booktabs}
\usepackage{amsmath}
\usepackage{amssymb}
\usepackage{enumitem}

\title{Correcting or Continuing?\\What On-Policy Distillation Learns from Student Errors}

\author{Shuo Yang\textsuperscript{1,2,3}~~~~ Zheyu Zhang\textsuperscript{1,3}~~~~ Mahito Sugiyama\textsuperscript{2}~~~~ Gjergji Kasneci\textsuperscript{1,3} \\
\\
 \textsuperscript{1}Technical University of Munich~~
 \textsuperscript{2}National Institute of Informatics
\\
  \textsuperscript{3}Munich Center for Machine Learning \\
  \small\texttt{shuo.yang@tum.de, zheyu.zhang@tum.de, mahito@nii.ac.jp, gjergji.kasneci@tum.de} \\}

\begin{document}
\maketitle

\begin{abstract}
On-policy distillation (OPD) trains a student on its own rollouts under a frozen teacher, mitigating the distribution mismatch of off-policy supervision. While prior work has primarily trained on problems a student can largely solve, we ask whether OPD can improve a student when training is restricted to problems it currently fails. We evaluate five representative OPD mechanisms with Qwen-3 and Phi-4 models under a unified competition-mathematics protocol. We find that OPD does not improve accuracy on this hard subset, with most mechanisms instead degrading performance despite continued decreases in training KL divergence. All eight configurations show a decline, and the two-sided sign test gives $p=0.0078$. In contrast, the same pipeline improves accuracy by 4.6 percentage points when training is not concentrated on failures, showing that the failure is conditional on the training regime rather than systematic. These findings suggest that OPD is effective when supervision is dominated by correct student prefixes, but can reinforce existing errors when training is concentrated on problems the student currently fails.
\footnote{Our code is available at \url{https://anonymous.4open.science/r/Correcting-or-Continuing-D74D/}.}
\end{abstract}

\section{Introduction}

Knowledge distillation aims to train a student model to imitate a stronger teacher model \citep{hinton2015distilling}. In on-policy distillation (OPD), the student generates its own outputs for a given task, and token-level supervision is computed from the Kullback–Leibler (KL) divergence between the teacher and student logits \citep{agarwal2024gkd,gu2024minillm}. This differs from earlier approaches that distill from a fixed corpus of teacher-generated samples rather than from the student's own rollouts \citep{kim2016sequence}. On-policy distillation has recently been proposed as an alternative to reinforcement learning with verifiable rewards for training small reasoning models \citep{thinkingmachines2025opd,ko2025distillm2,shao2024deepseekmath,deepseekai2025r1}. 


\begin{figure}[!t]
    \centering
       \includegraphics[width=\linewidth]{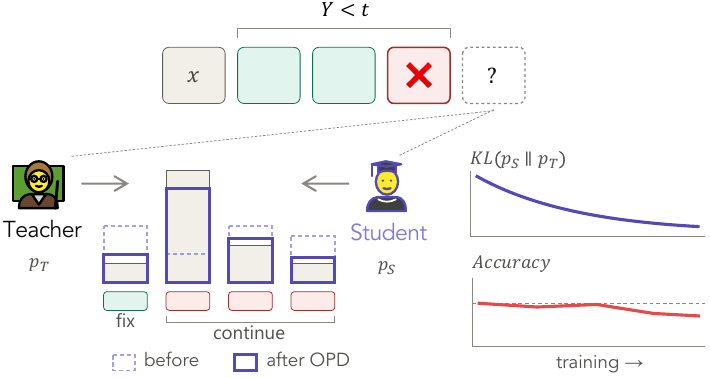}
    \caption{Closer to the teacher does not imply more correct. When the student’s prefix $y_{<t}$ already contains an error, on-policy distillation drives $p_S$ toward $p_T(\cdot \mid x, y_{<t})$, reducing KL divergence yet failing to improve accuracy relative to the frozen student.
    }
    \label{fig:example}
\end{figure}

Reported successes share a common regime in which supervision is dominated by correct student prefixes, either because the student is a tuned model or because much of the training distribution lies within its existing capabilities. We instead ask whether OPD can improve a student when training is restricted to problems it currently fails to solve, so that supervision is dominated by incorrect prefixes, as shown in Figure~\ref{fig:example}. Therefore, we evaluate five mechanisms under a unified experimental protocol: a trained prompt rewriter, a soft-prefix encoder, a residual-stream offset, a per-token cross-attention steering vector, and full fine-tuning. Furthermore, we consider both forward and reverse KL objectives, as well as direct imitation.
Our contributions are as follows:

\begin{itemize}[leftmargin=*, nolistsep]

\item \textbf{Characterizing when improvement remains possible.} We show that further gains are concentrated on problems the student currently fails to solve, suggesting that the remaining gap is primarily informational rather than a consequence of limited model capacity.

\item \textbf{Establishing a robust negative result.} Across five mechanisms, OPD fails to improve and often degrades accuracy, consistently across implementations with a sign test reaching $p=0.0078$.
\vspace{5pt}
\item \textbf{Identifying a conditional failure regime.} The same pipeline significantly improves a weaker student on an easier benchmark ($p=9.5\times10^{-4}$), demonstrating that the negative result depends on the student and task regime rather than the training pipeline itself.
\vspace{5pt}
\item \textbf{Tracing the failure.} Our analyses provide evidence that supervision from incorrect prefixes can reinforce existing errors, while direct imitation of reference solutions does not recover the lost performance. We further identify three evaluation artifacts that can produce spurious positive results in this setting.
\vspace{5pt}
\end{itemize}

Our findings suggest that OPD is effective when applied to a student whose rollouts are broadly correct, but can fail when supervision is concentrated on the narrow set of problems where the student already makes mistakes. This distinction becomes critical when applying OPD to already capable models.

\section{Problem Setup}
\label{sec:prelim}

We formulate reverse-KL on-policy distillation as minimizing the KL divergence from the student distribution to the teacher distribution along trajectories sampled from the student:
\begin{equation}
\label{eq:opd}
\mathbb{E}_{x,y\sim p_S(\cdot|x)}
\sum_t
\mathrm{KL}(
p_S(\cdot|x,y_{<t})
\|
p_T(\cdot|x,y_{<t})
).
\end{equation}
where $p_S$ denotes the current student policy and $p_T$ denotes a frozen teacher conditioned on the same prompt $x$ and student-generated prefix $y_{<t}$. For steering-based methods, the student backbone remains frozen, and $p_S$ is induced by composing the backbone with a trainable perturbation module, such as a soft prefix, a residual-stream offset, or a cross-attention steering vector.

The forward-KL variant uses the same on-policy sampling distribution, $y\sim p_S(\cdot\mid x)$, but reverses the arguments of the divergence:
\begin{equation}
\mathbb{E}_{x,y\sim p_T(\cdot|x)}
\sum_t
\mathrm{KL}(
p_T(\cdot|x,y_{<t})
\|
p_S(\cdot|x,y_{<t})
).
\end{equation}

Privileged self-distillation instead conditions the teacher on both the input $x$ and the reference solution $y^\star$, providing supervision from the correct solution path. Table~\ref{tab:opd-vs-opsd} in Appendix~\ref{app:setup} summarizes the distinction. Our main experiments use the unprivileged formulation, in which the teacher conditions only on the original input and the student's sampled prefix. We introduce reference-conditioned supervision only as a stress test in Section~\ref{sec:mechanism}.

\section{Experiment and Results}
\subsection{The Elicitation Ceiling}
\label{sec}

We first quantify how much external supervision could improve the student in principle. We define the elicitation ratio as
\begin{equation}
\label{eq}
\eta_{\mathrm{acc}} =
\frac{\mathrm{Acc}_{\mathrm{method}}-\mathrm{Acc}_{\mathrm{base}}}
{\mathrm{Acc}_{\mathrm{priv}}-\mathrm{Acc}_{\mathrm{base}}},
\end{equation}
where $\mathrm{Acc}_{\mathrm{priv}}$ is the accuracy obtained when the frozen student is provided with the full worked reference solution in context. We use the worked solution rather than an answer-only string because the former provides the information required to solve the problem, whereas the latter primarily measures answer copying. The denominator therefore represents the improvement available through direct access to the reference solution.

Because every mechanism in Section~\ref{sec:isolation} decreases rather than improves accuracy, $\eta_{\mathrm{acc}}$ is negative for all methods. More importantly, none of the methods recovers a positive fraction of the available elicitation ceiling. On 800 problems sampled from the training distribution, evaluated with thinking enabled and a 16k-token budget, the student solves 89.6\% without external supervision. As shown in Table~\ref{tab:eta}, 684 problems are therefore already solved by the student and provide no accuracy headroom. The remaining 116 problems contain nearly all of the available headroom: providing the reference solution increases accuracy by 26.9 percentage points, with a 95\% confidence interval of 21.3--33.0 percentage points. On the hardest bucket, the corresponding gain reaches 50 percentage points, compared with only 3.9 percentage points overall.

The substantial improvement from reference-conditioned supervision on the hardest problems indicates that the student retains sufficient capacity to solve these problems when provided with the relevant information. The remaining failure of OPD therefore points to a gap in eliciting existing capability rather than a lack of model capacity.

A second, independent measurement provides consistent evidence. Under a leakage-free constraint, clean prompt rewriting attains an accuracy ceiling of only about two percentage points. Furthermore, the teacher-likelihood reward used to select rewrites is negatively correlated with answer correctness, yielding a Spearman correlation of $\rho = -0.41$. This suggests that higher teacher likelihood does not reliably identify correct rewrites. Full details are provided in Appendix~\ref{app:setup}.

We therefore focus our OPD experiments on a hard subset of 1{,}034 problems selected from 8{,}000 candidates. Each selected problem has a base accuracy of at most 0.75. We split these problems into 904 training examples and 130 held-out examples.

\begin{table}[t]
\centering\small
\begin{tabular}{lrrrr}
\toprule
\textbf{Bucket (base acc.)} & $n$ & \textbf{Base} & \textbf{Priv.} & $\boldsymbol{\Delta}$ (pp) \\
\midrule
$= 0$             & 57  & 0.0   & 17.5  & +17.5 \\
$(0, 0.5)$        & 13  & 25.0  & 75.0  & +50.0 \\
$[0.5, 1)$        & 46  & 64.7  & 96.7  & +32.0 \\
$= 1$ (saturated) & 684 & 100.0 & 100.0 & 0.0 \\
\bottomrule
\end{tabular}
\caption{Headroom from the reference solution sits on problems the student fails.}
\label{tab:eta}
\end{table}

\subsection{subIsolation Study Across Mechanisms}
\label{sec:isolation}

\begin{figure}[t]
\centering
\includegraphics[width=\columnwidth]{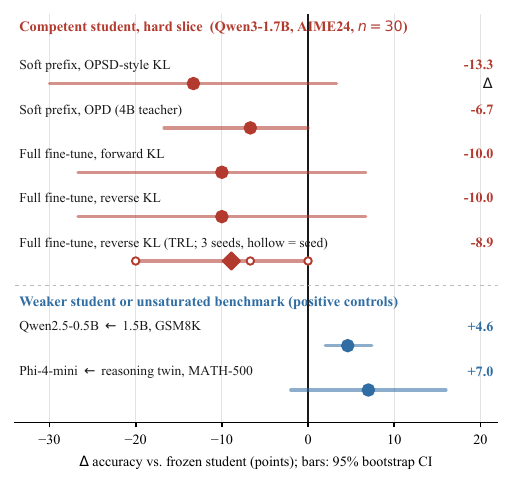}
\caption{Accuracy change relative to the frozen student on the competent-student hard subset. All evaluated mechanisms yield non-positive changes. Single-run results are shown except for the TRL configuration, for which three training seeds and their mean are reported. The same pipeline improves weaker students, as discussed in Section~\ref{sec:isolation}.}
\label{fig:summary}
\end{figure}

\textbf{Setup.} We distill Qwen3-1.7B-Instruct using a frozen Qwen3-4B-Instruct teacher~\citep{qwen2025qwen3} on the 904-problem hard subset. We vary the distillation mechanism, KL direction, and training implementation, comparing our in-house implementation with the standard TRL trainer. Figure~\ref{fig:summary} reports the change in AIME24~\citep{aime24} pass@1 accuracy relative to the untouched student. We use 30 problems, grade boxed answers with a math verifier, and allow up to 16k generated tokens.

A second, independently implemented codebase evaluates activation-level steering with the frozen student. It uses three mechanisms: a soft prefix, a residual-stream offset, and a per-token steering vector derived from cross-attention over the problem's hidden states. The corresponding results are reported in Table~\ref{tab:steering}. Appendix~\ref{app:iso} provides implementation details for all five mechanisms, Appendix~\ref{app:setup} reports the trained prompt-rewriter results, and Appendix~\ref{app:repro} provides the complete hyperparameter and hardware configuration.

\textbf{Results.} All evaluated configurations yield non-positive accuracy changes. Varying model capacity, KL direction, gating and warm-up, teacher scale, or training framework does not reverse this direction. Meanwhile, token-level KL decreases throughout training even when held-out accuracy declines.

We assess the consistency of this effect using a two-sided sign test \citep{dixonmood1946sign} under a symmetric null of zero effect. With five matched configurations, all five accuracy changes are negative, yielding a two-sided $p=0.063$ and a one-sided $p=0.031$. Including the three additional negative steering configurations in Table~\ref{tab:steering} gives eight negative signs out of eight and $p=0.0078$ (two-sided). We exclude the fourth steering configuration because its effect is consistent with recovery from an induced baseline deficit rather than a genuine capability change.

Individual configurations show substantially greater variability in magnitude. Across three TRL training seeds, the accuracy changes are $-6.7$, $-20.0$, and $0.0$ percentage points, with a mean of $-8.9$ and a standard deviation of $10.2$ points. None of the seeds yields a positive change. Re-evaluating the frozen student alone gives accuracies of 46.7\%, 40.0\%, and 40.0\%, with the sampling temperature of 0.6 changing the outcome on 10 of 30 AIME24 problems relative to greedy evaluation. These results motivate reporting the consistency of the direction rather than relying on the magnitude from any single run. Training seeds are treated as repeated measurements of the same configuration and are therefore not counted as independent observations in the sign test. Appendix~\ref{app:iso} provides further details.

\begin{table}[t]
\centering\small
\setlength{\tabcolsep}{2pt}
\begin{tabular}{@{}p{0.66\linewidth}p{0.14\linewidth}r@{}}
\toprule
\textbf{Configuration} & \textbf{Metric} & \textbf{$\Delta$} \\
\midrule
Residual offset / forward KL (8B) & Macro & $-2.2$ \\
\addlinespace[2pt]
\multicolumn{3}{@{}l}{\textit{Token attention}} \\
\hspace{1em}Sparse-accuracy RL (--) & Macro & $-11.1$ \\
\hspace{1em}Privileged imitation (4B) & AIME24 & $-11.7$ \\
\hspace{1em}OPD after imitation (4B) & AIME24 & $+1.7$ \\
\bottomrule
\end{tabular}
\caption{Independent steering results (avg@2). Teacher sizes are in parentheses and ``--'' denotes no teacher.}
\label{tab:steering}
\end{table}

\textbf{Positive controls.} To test whether the observed failure reflects a systematic issue in our implementation, we apply the same TRL reverse-KL pipeline to settings in which OPD has previously been shown to improve performance. Qwen2.5-0.5B-Instruct, distilled from Qwen2.5-1.5B-Instruct \citep{qwen2024technical} on GSM8K prompts \citep{cobbe2021training}, improves from 43.6\% to 48.2\% accuracy over 1{,}319 problems, corresponding to a statistically significant gain at $p=9.5\times10^{-4}$. Phi-4-mini-instruct \citep{abouelenin2025phi4mini}, distilled from a same-size Phi-4-mini-reasoning teacher \citep{xu2025phi4minireasoning}, improves from 67\% to 74\% on MATH-500 \citep{lightman2024letsverify}. This second control is not competence-matched and therefore serves only as an additional positive control. Full details are provided in Appendix~\ref{app:pc}.

These positive controls provide evidence against a general defect in the training pipeline. However, because they vary student capability, benchmark difficulty, and model family simultaneously, they do not isolate the factor responsible for the observed difference.

\subsection{Failure Mode: Supervision on an Incorrect Prefix}
\label{sec:mechanism}

Our setting differs from those in which OPD has previously succeeded: the student is already capable, receives no supervised warm-up, and is trained only on problems it currently fails. This creates a failure mode inherent to the on-policy objective in Equation~\ref{eq:opd}. The KL supervision is computed on the student's sampled prefix $y_{<t}$. When this prefix contains an incorrect step, the teacher is conditioned on an already erroneous state rather than on the correct solution path. The resulting supervision therefore cannot directly correct the error that generated the prefix and may instead reinforce the student's existing trajectory. Figure~\ref{fig:summary} is consistent with this mechanism: training KL decreases while held-out accuracy also declines.

\textbf{Stress test.} We ask whether the failure is simply caused by the absence of a gold solution as an explicit anchor. Providing supervision directly from the reference solution does not resolve the problem. Row three of Table~\ref{tab:steering} decreases accuracy by 11.7\% in a single run, which we interpret as evidence of direction rather than effect size given the seed variability above. A subsequent on-policy KL step recovers accuracy only to the untouched baseline.

A related soft-prefix analysis offers further evidence. Across 94 problems, the learned prefixes are nearly collinear, with pairwise cosine similarities of 0.998--0.999, whereas the underlying problem representations yield only 0.78 and a single direction accounts for 70--73\% of their variance. As the checkpoint corresponding to row three was not retained, these results constitute consistent rather than direct evidence. We therefore interpret the observed recovery as the removal of a self-induced deficit rather than the elicitation of new capability.

\textbf{Evaluation artifacts.} Three artifacts can produce spurious positive results in this setting. Permissive regex matching can overstate gains: an apparent 11.7\% improvement fell to 0.8\% after requiring a verified boxed answer. An insufficient generation budget can truncate reasoning and drive both baseline and treatment accuracy to 0, producing a misleading null result. Finally, answer leakage through unaudited rewrites can inflate apparent gains by 10--20 \%. Appendix~\ref{app:art} provides details.

\section{Conclusion}

Across five mechanisms and three training objectives, OPD does not reliably improve a student when training is restricted to problems it currently fails. The negative direction is consistent across eight configurations, while a positive control improves accuracy by 4.6 percentage points under the same pipeline, indicating that the failure depends on the training regime rather than on an implementation defect. These results delineate the boundary of standard OPD: it is effective when supervision is dominated by correct student prefixes, but can reinforce existing errors when supervision is concentrated on incorrect prefixes. 

Future work should isolate student capability from task difficulty under matched conditions, test whether outcome-based reinforcement learning exhibits a similar failure mode, and develop objectives that correct erroneous prefixes rather than imitate the teacher on them.

\section*{Limitations}
Our findings highlight the potential risks of applying on-policy distillation without carefully assessing its effects on model capabilities. In particular, training primarily on problems that the student cannot solve may reinforce erroneous reasoning rather than improve accuracy, potentially propagating errors to downstream applications. However, our negative evidence is drawn from Qwen3, and is evaluated primarily on one hard-mathematics subset. Its transferability to other settings therefore remains unknown. AIME24 contains only thirty items, therefore we accordingly claim directional consistency rather than effect magnitude throughout. 
Furthermore, training and evaluating multiple distillation configurations incurs computational costs and associated environmental impacts, although we do not quantify energy consumption in this work. These considerations underscore the importance of evaluating both task-level performance and error patterns before deploying models trained with on-policy distillation.

\section*{Ethical Considerations}

The findings of this work raise broader questions about responsibility in the development and evaluation of AI systems. Distillation can make a model appear to be learning from a more capable teacher, yet improvements in optimization metrics do not necessarily translate into greater practical competence. Treating such metrics as proxies for capability may create misplaced confidence in models whose errors remain unresolved. This is particularly consequential in settings where model outputs inform human judgments or decisions, as errors may be accepted or propagated when users overestimate a system's reliability. Our work highlights the importance of critically examining what training procedures actually teach models, rather than assuming that successful optimization implies meaningful capability gains. We propose that our results will motivate greater scrutiny of the assumptions underlying model evaluation and deployment.

\bibliography{custom}

\appendix

\section{Related Work}
\label{sec:related}

On-policy distillation and related generalized distillation objectives train a student on its own rollouts under the supervision of a frozen teacher \citep{agarwal2024gkd,gu2024minillm,ko2024distillm,ko2025distillm2}. Existing studies typically initialize the student from a base or weakly tuned model that has already received supervised warm-up \citep{thinkingmachines2025opd,qwen2025qwen3,yu2024metamath}. Recent work has identified several pathologies of on-policy supervision, including degraded supervision as the student's rollouts diverge from the teacher's distribution \citep{xie2026iwopd} and unreliable guidance when a generated prefix leaves the teacher's support \citep{fu2026revisitingopd}. These studies primarily diagnose such failures and propose modifications to the objective. In contrast, we investigate under what conditions the standard objective remains effective, evaluating multiple distillation mechanisms and validating our findings with both a positive control and a stress test.

Several recent extensions modify the standard distillation objective or its information available to the teacher. Reward-weighted generalizations can allow a student to surpass its teacher \citep{yang2026generalizedopd}. Our results concern the standard KL-based and imitation objectives on a single challenging domain and therefore do not directly address such reward-weighted formulations. Privileged self-distillation instead conditions the teacher on the reference solution \citep{zhao2026opsd}. We study the standard teacher-student setting without privileged information and consider reference-conditioned supervision only as a stress test.

Our experiments also include prompt rewriting as discrete forms of intervention \citep{yang2024opro,kong2024prewrite}. Our results highlight the need to examine rewriters for answer leakage when evaluating their effectiveness. Finally, our identified failure mode arises from supervision computed over the student's own sampled tokens, whereas reinforcement learning with verifiable rewards typically provides supervision from the final outcome \citep{shao2024deepseekmath,deepseekai2025r1}. Whether outcome-based reinforcement learning exhibits an analogous failure mode remains an open question, as does the broader question of whether such training induces genuinely new reasoning capabilities or primarily reinforces capabilities already present in the model's distribution \citep{yue2025rlvr,zhang2026consolidating}.

\section{Setup and Diagnostics}
\label{app:setup}

\begin{table}[ht]
\centering\small
\setlength{\tabcolsep}{2pt}
\begin{tabular}{@{}p{0.28\linewidth}p{0.27\linewidth}p{0.37\linewidth}@{}}
\toprule
\textbf{Aspect} & \textbf{OPD} & \textbf{OPSD} \\
\midrule
Teacher input & $x$ only & $x$ and $y^\star$ \\
Gold anchor & None & Reference path $y^\star$ \\
Positive gain & Distillation & Elicitation + write-in \\
\bottomrule
\end{tabular}
\caption{OPD lacks OPSD's privileged answer anchor, making its training problem harder.}
\label{tab:opd-vs-opsd}
\end{table}

\begin{figure}[ht]
\centering
\includegraphics[width=\columnwidth]{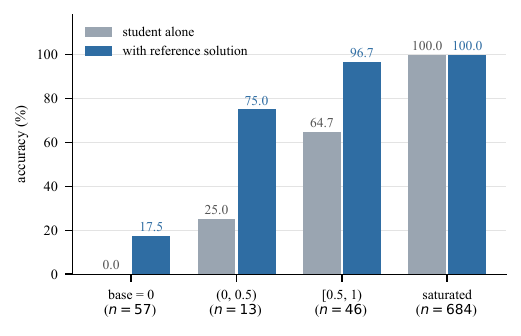}
\caption{Accuracy with and without the reference solution in context, by the student's own base accuracy, with thinking enabled and a 16k-token budget.}
\label{fig:ceiling}
\end{figure}

\textbf{Ceiling by rewriting.} Figure~\ref{fig:ceiling} plots the bucketed ceiling discussed. On MATH-500 we measure the accuracy a best-of-$K$ oracle rewriter can recover under a boundary that excludes any rewrite containing a solution or a leaked final answer. An initial bound of twenty points traces entirely to unterminated reasoning traces that leaked the answer. Once that boundary is enforced cleanly the bound falls to about two points, and permitting a short strategy hint without a solution raises it again to 8.5 to 10.5 points. No trained rewriter reproduced even this relaxed bound without either failing to update or re-leaking the answer. A KL-based reward used to select among rewrites is negatively correlated with downstream accuracy. We measure this with the Spearman rank coefficient \citep{spearman1904proof}
\begin{equation}
\label{eq:spearman}
\rho = 1 - \frac{6\sum_{i=1}^{n} d_i^2}{n(n^2-1)},
\end{equation}
where $d_i$ is the difference between the reward rank and the accuracy rank of rewrite $i$ among $n$ rewrites, giving $\rho=-0.41$ here, therefore, optimizing agreement with the teacher actively selects against correct rewrites rather than for them.

\textbf{Trained rewriters.} Under the clean boundary, a trained rewriter changed accuracy by only about one point, a result indistinguishable from noise, and a later rewriter checkpoint instead lowered AIME24 accuracy from 45.8\% to 26.7\% under four-sample averaging.

\textbf{Privileged context under a short budget.} Under a shorter, non-thinking budget the same reference solution becomes a net negative, since a long solution turns into interference once it can no longer fit the available context.

\section{Isolation Study Details}
\label{app:iso}

\begin{figure}[ht]
\centering
\includegraphics[width=\columnwidth]{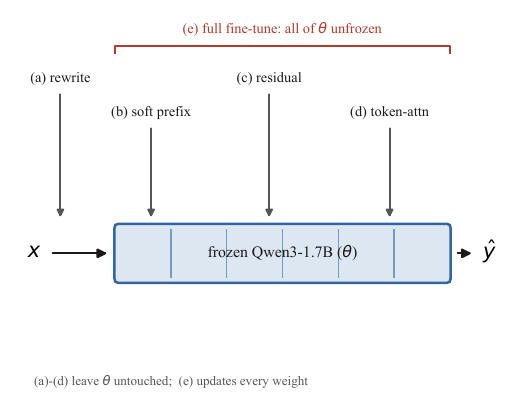}
\caption{The five mechanisms, ordered by the extent to which each modifies the frozen model. The first four leave all parameters of $\theta$ unchanged, whereas only full fine-tuning updates them.}
\label{fig:taxonomy}
\end{figure}

\begin{table}[ht]
\centering\footnotesize
\setlength{\tabcolsep}{3.2pt}
\resizebox{\columnwidth}{!}{
\begin{tabular}{@{}lrrrcr@{}}
\toprule
\textbf{Configuration} & \textbf{Base} & \textbf{Meth.} & $\Delta$ & \textbf{W/L/T} & $p$ \\
\midrule
Prefix, OPSD-style KL & 26.7 & 13.3 & $-13.3$ & 2/6/22 & 0.29 \\
Prefix, OPD (4B) & 26.7 & 20.0 & $-6.7$ & 0/2/28 & 0.50 \\
Full FT, forward KL & 40.0 & 30.0 & $-10.0$ & 2/5/23 & 0.45 \\
Full FT, reverse KL & 46.7 & 36.7 & $-10.0$ & 2/5/23 & 0.45 \\
TRL reverse KL, seed 42 & 46.7 & 40.0 & $-6.7$ & 3/5/22 & 0.73 \\
TRL reverse KL, seed 0 & 40.0 & 20.0 & $-20.0$ & 0/6/24 & 0.031 \\
TRL reverse KL, seed 1 & 40.0 & 40.0 & $0.0$ & 3/3/24 & 1.00 \\
\bottomrule
\end{tabular}}
\caption{AIME24 pass@1 accuracy with thirty problems for every configuration: accuracy in percent, delta, items won, lost, and tied by the method, and the exact two-sided McNemar $p$-value. Each configuration re-samples its own frozen baseline.}
\label{tab:cells}
\end{table}

\begin{figure}[ht]
\centering
\includegraphics[width=\columnwidth]{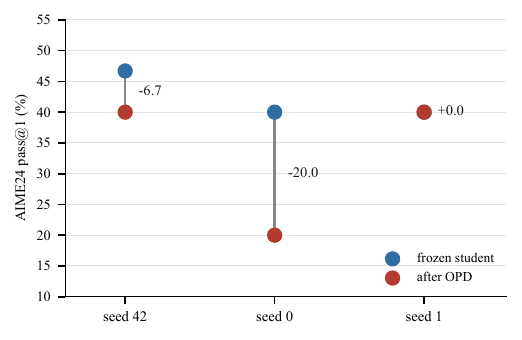}
\caption{Three training seeds of the TRL configuration. The frozen student itself varies between 40.0 and 46.7\% across evaluation runs, and its per-problem verdict changes on ten of thirty items.}
\label{fig:seed}
\end{figure}

\textbf{Sign tests.} Table~\ref{tab:cells} lists every cell in full, and Figure~\ref{fig:seed} plots the seed spread discussed in Section~\ref{sec:isolation}. The sign test treats each configuration's sign as a fair coin under a null of no effect: five negative outcomes out of five gives a one-sided $p=0.031$, or $0.063$ two-sided. Adding the three negative steering rows gives eight out of eight, at a two-sided $p=0.0078$. We exclude row four of Table~\ref{tab:steering} throughout, since we read it as recovery from an induced deficit rather than as an independent configuration. All configurations share the same thirty items, thus this test establishes sign consistency across recipes rather than independence across 150 separate problems. Seeds and later re-evaluations re-measure configurations that are already counted and are accordingly never added a second time. No single item is degraded by all five configurations at once.

\textbf{Per-cell significance.} For each cell we also report an exact McNemar test \citep{mcnemar1947note} on the items where the method and the frozen student disagree. Writing $b$ for the number of items the method alone solves and $c$ for the number the base model alone solves, with $n=b+c$,
\begin{equation}
\label{eq:mcnemar}
p_{\mathrm{two}} = \min\!\left(1,\ 2\sum_{i=0}^{\min(b,c)} \binom{n}{i} 2^{-n}\right)
\end{equation}
is the exact two-sided $p$-value under a null that each disagreement is equally likely to favor either side. This is the test behind every $p$ reported in Table~\ref{tab:cells}. Confidence intervals instead use a nonparametric bootstrap \citep{efron1979bootstrap}: resampling the $n$ graded items with replacement $B$ times, computing the mean paired delta $\Delta^{*}_b=\tfrac{1}{n}\sum_{i} (y^{\mathrm{method}}_{i} - y^{\mathrm{base}}_{i})$ on each resample, and reporting the 2.5th and 97.5th percentiles of $\{\Delta^{*}_b\}_{b=1}^{B}$ with $B=10{,}000$.

\textbf{AIME25 re-evaluation.} Re-evaluating the two full fine-tuning checkpoints on AIME25~\citep{aime25} under the identical protocol gives a delta of $+3.3$ points for both, each a single additional solved item at $p=1.0$, the opposite sign of their AIME24 deltas of $-10.0$ and $-6.7$ points. Pooling both benchmarks for a sample of sixty returns the direction to negative, at $-3.3$ and $-1.7$ points respectively, again without significance. We read this as a concrete illustration of the single-benchmark noise floor rather than as evidence against the overall direction, and we do not add these evaluations to the sign test above.

\begin{figure}[ht]
\centering
\includegraphics[width=\columnwidth]{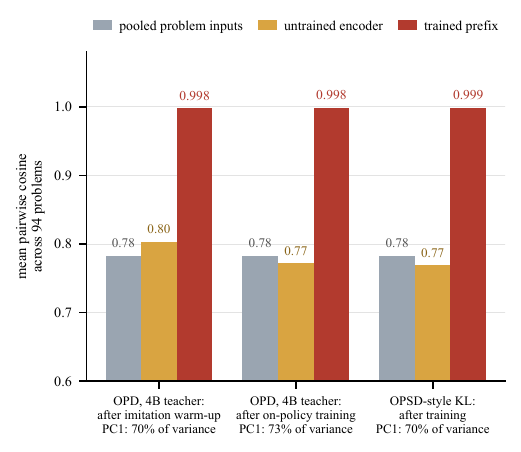}
\caption{Trained soft prefixes are nearly collinear across ninety-four problems, drawn from all thirty AIME24 items plus sixty-four further hard-subset items, unlike the pooled problem inputs or an untrained encoder of identical architecture. Static embeddings serve as a proxy input for the checkpoints trained with contextual pooling. A high cosine under this proxy remains informative even so.}
\label{fig:collapse}
\end{figure}

\textbf{Prefix collapse.} Across these ninety-four problems the trained prefixes are nearly collinear, as Figure~\ref{fig:collapse} shows, and the update from the imitation phase to the on-policy phase is itself a shared shift rather than a per-problem correction: the pairwise cosine between these shift vectors is 0.999. The learned prefix thus behaves as a global offset rather than as an instance-conditioned intervention.

\textbf{Steering codebase.} Figure~\ref{fig:taxonomy} shows where each mechanism attaches to the frozen backbone. A small encoder maps each problem to a soft prefix, to a single offset added to every prompt embedding, or to a per-token offset produced by cross-attention over hidden states of the problem's own tokens, while the student itself stays entirely frozen throughout. The privileged-imitation row of Table~\ref{tab:steering} trains this offset with cross-entropy against the reference solution directly, and the following row applies a further on-policy KL step on top of it.

\section{Positive Controls}
\label{app:pc}

\begin{table}[ht]
\centering\footnotesize
\resizebox{\columnwidth}{!}{
\begin{tabular}{@{}lrrrrr@{}}
\toprule
\textbf{Control} & \textbf{Base} & \textbf{Student} & $\Delta$ & \textbf{$n$} & $p$ \\
\midrule
Qwen2.5-0.5B, GSM8K & 43.6 & 48.2 & $+4.6$ & 1319 & $9.5{\times}10^{-4}$ \\
Phi-4-mini, MATH-500 & 67.0 & 74.0 & $+7.0$ & 100 & 0.21 \\
Phi-4-mini, AIME24 & 10.0 & 13.3 & $+3.3$ & 30 & 1.00 \\
\bottomrule
\end{tabular}}
\caption{Positive controls under the identical TRL reverse-KL pipeline. For GSM8K the student wins 196 items and loses 135, with a 95\% confidence interval of 2.0 to 7.4 points. For Phi-4-mini on MATH-500 the student wins 15 items and loses 8, with a 95\% confidence interval of $-2.0$ to 16.0 points.}
\label{tab:pc}
\end{table}

\textbf{GSM8K.} Table~\ref{tab:pc} summarizes both controls in full. We train with TRL reverse KL at $\beta=1$, on prompt-only data drawn from the 7{,}473 GSM8K \citep{cobbe2021training} training problems, for one epoch, under an 8k generation cap, and we evaluate greedily on the 1{,}319 test problems.

\textbf{Phi-4-mini.} This tokenizer-family control distills Phi-4-mini-instruct \citep{abouelenin2025phi4mini} from Phi-4-mini-reasoning \citep{xu2025phi4minireasoning}, both 3.8B parameters and sharing a vocabulary, on 800 problems from the MATH training set \citep{hendrycks2021measuring} for 200 steps. This control does not constitute a competence-matched replication of our main setting. The teacher is of the same size as the student rather than larger, MATH-500 is initially unsaturated for this particular student, and most rollouts were truncated at the 8k cap rather than terminating naturally. The MATH-500 gain does not reach statistical significance, and the AIME24 gain reduces to a single additional item. We therefore interpret this control only as a directional signal that the observed pattern is not specific to Qwen3.

\begin{table*}[ht]
\centering\footnotesize

\begin{tabular}{lcccc}
\toprule
 & \textbf{Isolation} & \textbf{Seed repeats} & \textbf{GSM8K} & \textbf{Phi-4} \\
\midrule
Student & Qwen3-1.7B & Qwen3-1.7B & Qwen2.5-0.5B & Phi-4-mini \\
Teacher & Qwen3-4B & Qwen3-4B & Qwen2.5-1.5B & Phi-4-reason. \\
Train set & hard-904 & hard-904 & GSM8K-7473 & MATH-800 \\
Steps run & 200 & 200 & 934, 1 epoch & 200 \\
Learning rate & 1.0e-5 & 1.0e-5 & 2.0e-5 & 1.0e-5 \\
Batch $\times$ accum. & 1 $\times$ 4 & 1 $\times$ 4 & 2 $\times$ 4 & 1 $\times$ 4 \\
Warmup steps & 20 & 20 & 28 & 10 \\
Train sampling & 0.6/0.95/20 & 0.6/0.95/20 & 0.6/0.95/20 & 0.6/0.95/20 \\
Eval decoding & sampled & sampled & greedy & greedy \\
Gen. cap, train & 16384 & 16384 & 8192 & 8192 \\
Thinking mode & on & on & off & off \\
Attention impl. & sdpa & sdpa & sdpa & eager \\
Seed & 42 & 0, 1 & 42 & 42 \\
\bottomrule
\end{tabular}
\caption{Hyperparameters shared by every TRL \texttt{DistillationTrainer} run in the paper, reverse KL at $\beta=1$, weight decay 0.01, gradient clip 1.0, the AdamW optimizer \citep{loshchilov2019decoupled} with $\beta_1=0.9$, $\beta_2=0.999$, $\epsilon=10^{-8}$, and a linear learning-rate decay. Sampled evaluation uses temperature 0.6 with top-$p$ 0.95 and top-$k$ 20.}
\label{tab:repro-trl}
\end{table*}

\begin{table*}[ht]
\centering\footnotesize
\begin{tabular}{lcccc}
\toprule
 & \textbf{Fwd KL} & \textbf{Rev KL} & \textbf{Prefix, OPD} & \textbf{Prefix, OPSD} \\
\midrule
Updates & full params & full params & prefix, 64 tok. & prefix, 96 tok. \\
Imitation warm-up & none & none & 300 steps & 500 steps \\
KL direction & forward & reverse & forward & OPSD-style \\
KL support & top-64, clip 5 & top-64, clip 5 & top-64, clip 5 & top-64, clip 5 \\
Train rollout cap & 16384 & 16384 & 4096 & 4096 \\
Steps run, evaluated & 400 & 800, at 500 & 400 & 400 \\
Batch $\times$ accum. & 1 $\times$ 1 & 4 $\times$ 1 & 4 $\times$ 1 & 1 $\times$ 1 \\
Learning rate & 1.0e-5 & 1.0e-5 & 2.5e-5 & 2.5e-5 \\
Weight decay, clip & 0.01, 1.0 & 0.01, 1.0 & 0.01, 0.5 & 0.01, 0.5 \\
Gated on errors & on & off & on & n/a \\
\bottomrule
\end{tabular}
\caption{Hyperparameters for the in-house engine behind the remaining four rows of Table~\ref{tab:cells}, all with a constant learning rate, no warmup, and the AdamW optimizer \citep{loshchilov2019decoupled}. Evaluation always uses a 16384-token cap regardless of the training rollout cap above. A shorter training cap is a property of the training setup only.}
\label{tab:repro-house}
\end{table*}

\section{Evaluation Artifacts}
\label{app:art}

Under regex extraction of the last number within a short generation budget, an apparent AIME24 gain of 11.7 percentage points, from 14.2\% to 25.8\%, decreased to merely 0.8 percentage points, from 18.3\% to 19.2\%, once boxed-answer verification was instead required on the same checkpoint. A 2k-token budget truncates the student's reasoning before it reaches a final answer. Consequently, under this budget the frozen baseline itself scores zero, which can be readily mistaken for a genuine effect if the baseline is not evaluated under the same budget. Finally, an unaudited rewriter that generated unterminated reasoning traces or stray boxed answers in its rewrites inflated apparent gains by ten percentage points under imitation and by twenty percentage points under oracle selection. Both apparent gains disappeared once the leakage boundary was properly enforced.

\section{Reproducibility}
\label{app:repro}

This appendix lists the hardware, software, and training hyperparameters behind every result in the paper.

\textbf{Hardware and software.} Every training and evaluation run uses a single compute node with one NVIDIA H200 NVL GPU, 141 GB of device memory, an AMD EPYC 9654 CPU with 96 cores, and 755 GiB of system memory, running NVIDIA driver 580.178.04 with system CUDA 12.9. The student and teacher always share this one GPU. The software stack is Python 3.11, PyTorch 2.10.0 built against CUDA 12.8, Transformers 4.57.6 \citep{wolf2020transformers}, TRL 1.13.0, and vLLM 0.18.0 \citep{kwon2023efficient} for sampling-heavy evaluation. All training uses bf16 mixed precision \citep{micikevicius2018mixed} with gradient checkpointing \citep{chen2016training}. Table~\ref{tab:repro-trl} lists hyperparameters for every TRL-trained run, and Table~\ref{tab:repro-house} for the in-house engine behind the remaining four isolation-study rows.

\end{document}